\documentclass[runningheads]{llncs}
\usepackage{hyperref}
\usepackage{amsmath}
\usepackage{float}
\usepackage{afterpage}
\usepackage{booktabs}
\usepackage{graphicx}
\usepackage{microtype}
\usepackage{makecell}
\usepackage{parskip}
\usepackage{etoolbox}
\usepackage{url}
\apptocmd{\thebibliography}{\raggedright}{}{}
\begin{document}
\title {Privacy-Preserving Personalized Fitness Recommender System \textbf{\texorpdfstring{$(P^{3}FitRec)$}{(P3FitRec)}}: A Multi-level Deep Learning Approach}

\titlerunning{Privacy-Preserving Personalized Fitness Recommender System \textbf{\texorpdfstring{$(P^{3}FitRec)$}{TEXT}}}
%
\author{Xiao Liu* \and
Bonan Gao* \and
Basem Suleiman$^{\dagger}$ \and
Han You* \and
Zisu Ma* \and
Yu Liu* \and
Ali Anaissi$^{\dagger}$}
\authorrunning{Liu and Gao, et al.}
%
\institute{School of Computer Science, The University of Sydney, Australia\\
$^{*}$\email{\{xliu7788, bgao9725, hyou8279, zima9310, yliu8248\}@uni.sydney.edu.au}\\
$^{\dagger}$\email{\{basem.suleiman, ali.anaissi\}@sydney.edu.au}}
\maketitle              
\begin{abstract}
Recommender systems have been successfully used in many domains with the help of machine learning algorithms. However, such applications tend to use multi-dimensional user data, which has raised widespread concerns about the breach of users' privacy. Meanwhile, wearable technologies have enabled users to collect fitness-related data through embedded sensors to monitor their conditions or achieve personalized fitness goals. In this paper, we propose a novel privacy-aware personalized fitness recommender system. We introduce a multi-level deep learning framework that learns important features from a large-scale real fitness dataset that is collected from wearable IoT devices to derive intelligent fitness recommendations. Unlike most existing approaches, our approach achieves personalization by inferring the fitness characteristics of users from sensory data and thus minimizing the need for explicitly collecting user identity or biometric information, such as name, age, height, weight. In particular, our proposed models and algorithms predict (a) personalized exercise distance recommendations to help users to achieve target calories, (b) personalized speed sequence recommendations to adjust exercise speed given the nature of the exercise and the chosen route, and (c) personalized heart rate sequence to guide the user of the potential health status for future exercises. Our experimental evaluation on a real-world Fitbit dataset demonstrated high accuracy in predicting exercise distance, speed sequence, and heart rate sequence compared to similar studies. Furthermore, our approach is novel compared to existing studies as it does not require collecting and using users' sensitive information, and thus it preserves the users' privacy.

\end{abstract}
\section{Introduction}
\label{sec:introduction}
\raggedbottom
With the advent of artificial intelligence, many innovative enterprises have provided various personalized products and services to individuals. Personalization is generally made possible by mining multidimensional data from individuals so that the products and services can be custom-made to people’s inclinations. In particular, the research of personalization on improving individuals’ well-being has attracted a lot of attention. For instance, smartphones and smartwatches can sense the dynamic changes of users, such as heart rate, blood pressure, sleep patterns, etc., to realize real-time and non-invasive monitoring of human health \cite{s18061714}. A study found that smartwatches can effectively monitor the instantaneous changes in the user's heart rate, and can accurately help diagnose atrial fibrillation associated with ischemic stroke (asymptomatic or paroxysmal) \cite{raja_elsakr_roman_cave_pour-ghaz_nanda_maturana_khouzam_2019}. On the other hand, other products focus on providing personalized suggestions to users, such as customizing exercise plans for users, understanding their reactions to the plans, and analyzing exercise results to continuously improve user experiences. For example, a framework named PRO-Fit (Personalized Recommender and Organizer Fitness assistant) was introduced, which uses collaborative filtering on user profile and activity data to generate personalized fitness schedules according to user availability and fitness goals \cite{10.1007/s00779-017-1039-8}. Traditionally, people can achieve comparative goals by hiring a human trainer, but machine learning-based applications have certain advantages in terms of quality and affordability. More specifically, machine learning algorithms can integrate various information sources from thousands or even millions of users to develop products or services, which is far beyond the knowledge range of a human trainer. In addition, recommender systems as software products usually have a lower marginal cost to provide services to new users.
However, despite the convenience brought by technology, societies are paying more attention to the security and privacy issues in data mining and prediction. In 2018, the European Union issued General Data Protection Regulation (GDPR) to regulate the protection of citizens' personal data and privacy. In this sense, Tan et al. \cite{5772388} and Bouhenguel et al. \cite{4810232} discussed the Bluetooth security threats of wearable devices, and share their knowledge and insights on how to prevent devices and the networks they are connected to from being attacked. Other researchers such as Cyr et al. \cite{Cyr2014SecurityAO} analyzed the potential security problems of Fitbit devices when collecting and utilizing users' data, such as unnecessarily collecting information from nearby devices, or withholding all collected data to the device owners. Unfortunately, most service or product providers are inherently motivated to collect as much data as possible, especially personal data, from users to cultivate machine or deep learning models to enhance user experience. 

Although most of the existing work in the literature focuses on building accurate artificial intelligence models to make personalized fitness recommendations, little consideration is given to protecting the privacy of users. For instance, Dharia et al. \cite{10.1007/s00779-017-1039-8} proposed a PRO-Fit framework to proactively push notifications recommending fitness-related activities to users, which is based on their multivariate data, including their fitness preferences, calendar data, and social network data. Although the model has achieved good performance, the framework inevitably collects an intensive amount of sensitive data from users. Likewise, the “TweetFit” framework which was proposed in \cite{Farseev_Chua_2017} profiles people’s wellness by taking advantage of data from sensors such as speed or heart rate measurements during exercises and multiple social media sources such as Twitter tweets and Instagram image captions and comments. On the other hand, with specific attention to GDPR compliance, Sanchez et al. \cite{Sanchez2020} presented a fitness data privacy model that learns people’s privacy preferences for fitness data sharing and processing collected by the Internet of things (IoT) devices. They studied user privacy permission settings of fitness trackers such as Fitbit and smartwatches with the supplement of a questionnaire to learn users’ privacy profiles by applying machine learning modeling. They further developed a rule-based personal data manager (PDM) framework to provide privacy advice to users based on their machine learning models. However, this work mainly focuses on the general privacy settings of IoT devices, not their product functions. Loepp et al.\cite{ubo_mods_00115757} proposed a prototypical smartphone app that recommends personalized running routes based on analyzing multivariate data of the workout routes users have run, e.g., length of the route, uniqueness of the route, the shape of the route, the light of the route, and so on. The framework makes personalized recommendations based on running routes following a rule-based approach and users’ preferences which were manually set by users. Compared with the rule-based approach, machine or deep learning-based recommender systems have the advantage of learning user attributes intelligently and require minimum manual input from the users.

In this paper, we address the above challenges by proposing a novel privacy-aware personalized fitness recommender system. In our proposed approach, we introduce a multi-level deep learning framework that learns important features from a large-scale real fitness dataset collected from wearable IoT devices to derive intelligent fitness recommendations. Unlike most existing approaches, our approach achieves personalization by inferring the fitness characteristics of users from sensory data and thus minimizing the need for explicitly collecting user identity or biometric information, such as name, age, height, and weight.

Our proposed approach consists of two key components. We first build a model that learns user embedding and workout route embedding from a real-world Fitbit dataset. The user and workout route embeddings are then fed as input features to our proposed deep learning models to create personalized recommendations. Second, we develop a workout profile prediction model that suggests personalized workout recommendations that can guide the user based on their choice of workout distance, route and speed sequence, sport type, and target calories to consume. The goal of our exercise distance prediction is to provide personalized guidance for users to achieve the target exercise calorie. Similarly, our predicted speed sequence aims to guide the user to adjust the exercise speed according to the conditions of the selected target exercise calories and the selected exercise route. Meanwhile, our predicted heart rate sequence aims to provide the user with an important indicator of the expected health status in the upcoming exercise.

In our approach, we also propose a three-dimensional tensor of "users – workout routes – contextual features" based on the historical workout data. The goal of this tensor is to capture the underlying structures inherited in users and workout routes using the Tensor Decomposition method CANDECOMP/PARAFAC (CP) \cite{kolda2009tensor}. The two resultant matrices related to the latent characteristics of the users and workout routes are then combined with other contextual features (choice of sport, target calories, etc.) as input to two models. The first model is a Multi-Layer Perceptron (MLP) which is used to predict the total distance of a future workout. The second model is a Long Short-Term Memory (LSTM) which is used to predict the speed sequence and heart rate sequence of a future workout.

The main contributions of our proposed approach are threefold:

\begin{enumerate}
\item An approach to building privacy-aware personalized fitness recommendation systems, by inferring the fitness characteristics of users from sensory data instead of collecting multidimensional private data from users. This is complementary to the personalized fitness recommendation systems that do not consider privacy preservation, and the privacy preservation approaches running on independent encryption protocols.
\item A model that learns user embedding and workout route embedding from real workout datasets collected from Fitbit devices. This includes gender, sport type, calories, workout duration, workout distance, workout speed, workout heart rate, and workout route geographical data. Our user and workout route embeddings are further utilized as the input features for our proposed privacy-aware and personalized fitness recommender.
\item A workout profile prediction model that suggests personalized recommendation of the necessary workout distance, the rhythm of speed, the change of heart rate of a future exercise based on the user’s choice of workout route, sport type, and target calories to consume.
\end{enumerate}

The rest of this paper is structured as follows. In section \ref{sec:related work}, we present background information relevant to the topics of our proposed approach. We also, discuss various related studies in the literature and practice.
The details of our proposed approach, including our methods and algorithms, are introduced in section \ref{sec:method}. The experimental evaluation and result analysis are discussed in section \ref{sec:evaluation}. In section \ref{sec:discussion}, we discuss the results in the context of our research goals and contributions. We draw key conclusions and future work in section \ref{sec:conclusion}.  
\section{Related Work}
\label{sec:related work}

In this section, we first briefly review the technologies for sequential data modeling and technologies used in building recommender systems. Then we discuss related studies on different fitness recommendation topics, e.g., Fitness Activity Detection and Recommendation, Sequential Fitness Profile Recommendation, and Privacy Preservation in Fitness Recommendation.

\textbf{Recurrent Neural Networks (RNNs).} In recent years, RNNs and particularly LSTM networks have been widely used in processing time-series data for sequential modeling tasks, for example, speech recognition \cite{Jorge_2019}, machine translation \cite{NIPS2014_a14ac55a}, image captioning \cite{Chu2020}, etc. Ilya Sutskever et al. proposed an end-to-end approach to address sequence to sequence mapping problem by constructing a multi-layer LSTM network, which maps the input sequence to a fixed-dimensional vector through LSTM, and then uses another LSTM to output the target sequence \cite{NIPS2014_a14ac55a}. They put forward several innovative techniques to improve the model, such as using stacked LSTM structure and reversing input vectors, etc. In English-French translation tasks, their model shows better performance over single forward LSTM models. Moreover, Bi-directional LSTM is an extension of traditional LSTMs that improves model performance on many sequential modeling tasks. This is because a unidirectional LSTM network only propagates information from past to future, while a Bi-directional LSTM network manages inputs in two ways, one from past to future and one from future to past \cite{650093}. Therefore, it consolidates the context of each step in the input sequence. Similarly, in our work, we consider a Bi-directional stacked LSTM architecture to solve the sequence-to-sequence modeling task.

\textbf{Technologies of Recommender Systems.} Recommender systems generally follow two basic approaches: collaborative filtering or content-filtering \cite{10.1145/2481244.2481250}. Collaborative filtering holds the belief that people would continue the same experience in the future if they liked something in the past. The K-nearest Neighboring (KNN) method is often used to be the most favored approach to conducting collaborative filtering, which conducts finding similar users’ profiles to one certain user to compute likeness and dislikes for an item \cite{10.1145/371920.372071}. In contrast, content-based filtering methods are useful in situations where user information is sufficient but not item information. It works as a classifier to model the likes and dislikes of the users to evaluate an item \cite{10.1145/371920.372071}. In collaborative filtering-based recommender systems, a two-dimensional data matrix associated with users and items is usually constructed. The 2-D matrix can be factorized into two matrices, namely a user matrix and an item matrix that contain the latent characteristics describing the user preferences and item profiles \cite{10.1145/1864708.1864727}. One disadvantage of the matrix factorization approach is that the context is only a scalar, representing the user's rating of the item. To overcome this shortcoming, Karatzoglou et al. discussed a multidimensional equivalent method to the 2-D matrix factorization approach named Tensor Decomposition \cite{10.1145/1864708.1864727}. Contrary to the single rating feature in the matrix decomposition method, a multidimensional contextual feature vector is established between each user and item. This method allows flexible integration of context information when learning entity embeddings to provide context-aware recommendations. In our work, we consider establishing a rich context feature vector between each user and workout route to derive user embeddings and workout route embeddings.

\textbf{Fitness Activity Detection \& Recommendation.} Guo et al. proposed FitCoach, which is a virtual fitness coach built upon data sensed by IoT devices \cite{8057208}. It aims at detecting people’s workout statistics such as exercise types with a lightweight support vector machine (SVM) classifier and providing fine-grained feedback on exercise form scores, i.e., motion strength and performing period, to assist users to maintain proper exercise postures and avoid injuries. Similarly, Zhao et al. introduced a fitness recommender system designed to generate personalized and gamified content to promote daily physical activities \cite{pmid33200994}. They collected various types of user data and built separate sub-models for user profile prediction with a non-machine learning approach. The sub-models work individually, but their results are jointly input into a decision tree-based recommendation engine to create personalized recommendations, such as extending an existing exercise or suggesting a different type of activity. Yong et al. proposed an IoT-based intelligent fitness system that monitors people’s health with data collected by IoT devices, recognizes people’s actions using a convolutional neural network (CNN) based model, and provides fitness-related recommendations, such as reminding users to attend fitness courses or going to gyms based on user predefined exercise plans \cite{10.1016/j.jpdc.2017.05.006}. They explicitly collected users’ scores on the exercise items to build a collaborative filtering based recommender system to realize personalization. Similarly, Saumil Dharia et al. presented the PRO-Fit framework that collects users' multivariate data, including fitness preferences, calendar data, and social network data \cite{10.1007/s00779-017-1039-8}. It applies machine learning algorithms to classify users’ activities into specific types, which are then used to establish user profiles reflecting their current lifestyle (sedentary vs. active). These user profiles are further fed into a collaborative filtering-based recommender system for personalized fitness activity or fitness partner recommendations. Unlike most studies that target the general public, Mogaveera et al. introduced a health monitoring and fitness recommendation system using machine learning targeting patients \cite{9358605}. The system collects data from both patients (body details, disease \& health records) and doctors (disease categories) to monitor patients’ condition based on some predefined rules and to provide personalized recommendations of diet and exercise plans through a decision tree based machine learning model. 

Compared with the existing research, our focus is on recommending dynamic changes of workout speed and heart rate of an exercise, which are complementary dimensions in the field of fitness recommendation. In addition, the construction of these exercise activity recommendation systems usually boils down to recommending some predefined fitness categories. In contrast, we propose a deep learning framework to solve a rarely studied sequence-to-sequence regression task to model workout speed and heart rate. Moreover, as discussed above, most researchers have achieved personalization with rule-based or collaborative filtering-based methods. In contrast, we propose learning user profiles based on tensor decomposition. Compared with the rule-based method, this method is considered to be more scalable and compared with the collaborative filtering method, it can realize context-aware user profiles \cite{10.1145/1864708.1864727}. 

\textbf{Sequential Fitness Profile Recommendation.} Berndsen et al. proposed a recommender system that predicts the target finish time of Marathon with an XGBoost model, followed by a collaborative filtering \&  K-Nearest-Neighbours (KNN) based framework to generate pacing recommendations for runners to achieve the target finish time \cite{9356261}. More specifically, historical training data of athletes, such as GPS data, completion time, and pacing information, are used to predict the target completion time. Meanwhile, a runner’s user profile can be inferred by applying collaborative filtering and further used to find the successful marathon finishers with the most similar user profiles applying K-Nearest-Neighbours (KNN) algorithm. Finally, pacing strategy recommendation is achieved by using the pacing strategies from these successful marathon finishers with similar user profiles. Compared with their work, the exercises we focus on are less competitive than Marathon, such as jogging, biking, etc. However, our task of predicting workout distance is similar to the prediction of target finish time in \cite{9356261}, except that we propose a multi-layer perceptron model to solve this task. In addition, their approach to workout pacing recommendation is through looking up historical exercises and following existing pacing records. We propose an LSTM based model to generate personalized new pacing sequences. 

Jianmo et al. proposed a personalized fitness recommendation system named FitRec to solve two tasks: predicting the heart rate and speed at all time steps of a future workout, and short-term prediction of heart rate and speed at a specific time step during an ongoing workout \cite{10.1145/3308558.3313643}. The former predicts a future workout profile to show users the anticipated performances in terms of heart rate and speed while the latter predicts transient heart rate and speed during an ongoing exercise, so as to facilitate tasks like anomaly detection or real-time decision-making. FitRec was primarily designed based on LSTM modules. More specifically, they used LSTMs to project the sequence of the most recent workout measurements of the user into a dense vector, which forms the user embedding vector. Then, the user embedding vector is concatenated with other input attributes of the new workout sequence as the combined contextual sequence input for two different LSTM networks corresponding to the two tasks and adopt a 2-layer stacked LSTM architecture and an LSTM-based encoder-decoder architecture respectively. 

Compared with FitRec, our work differs in many ways, although we study the same dataset as FitRec, and its task of predicting a future workout profile has a similar purpose to ours. First, we suggest adding caloric expenditure to the input features, so that our system can provide different recommendations according to users’ inputs of target caloric expenditure, which enhances the practical value of our research. Furthermore, FitRec predicts the speed of a workout using distance and time as inputs, which arguably may be seen as a trivial issue solvable by simple math. Therefore, we propose removing the time sequence feature from the inputs to avoid potential information leakage. Besides, we improve the derivation of entity embeddings by learning them from all historical workout records of the users, instead of learning only from the latest one. Our tensor decomposition method for learning entity embedding is also different from theirs.

\textbf{Privacy Preservation in Fitness Recommendation.} In privacy-aware recommendation systems, privacy preservation is usually achieved through some independent encryption protocols in the process of data collection or data exchange \cite{Sanchez2020}. \cite{BEG2021102874} proposed a reversible data transform algorithm based privacy-preserving data collection protocol for mobile app recommendation systems. According to this protocol, a user's data is sent through a user group with encryption, which avoids direct communication between a user and a data collector. Likewise,  Arijit Ukil addresses the privacy preservation problem through a random security key pre-distribution method \cite{5628748}. According to the proposed scheme, private data can be collected from various sources and aggregated by the service provider or the server securely. \cite{4664769} studies the challenge of privacy preservation in a more specific scenario when the recommendation systems of two independent business entities are merged. In this scenario, both a homomorphic encryption approach and a scalar product approach are proposed to encrypt raw data before data exchange takes place between the two systems. Badsha et al. also proposed a homomorphic encryption approach to enforce privacy preservation in building a recommender system \cite{Badsha2016}. More specifically, they collected encrypted ratings on items from users and sent ciphertexts to recommender servers to calculate the similarity among the rated items homomorphically. The similarity scores were subsequently decrypted by the users without revealing any private information and then used to build a recommender system by using content-based filtering and collaborative filtering methods.

In contrast, our work focuses on developing the recommendation algorithm itself with minimum private data, which is complementary to the data encryption approach discussed above. Sanchez et al. conducted a study on users' preferences on privacy permission in the fitness domain and recommended a series of strategies for users to set permissions according to the collected and shared data \cite{Sanchez2020}. They found that users have the highest acceptance rates of privacy permissions in terms of gender and fitness types, while the users are reluctant to share their height, weight, age, and social network information. The results of this study confirm our hypothesis about privacy protection and are consistent with our attempt to construct a personalized fitness recommendation system by using minimum private information from the users. To the best of our knowledge, most existing work in the fitness recommendation domain attempted to collect multidimensional demographic parameters from the users to improve the performances of their recommendation algorithms. In this paper, we demonstrate that a recommendation algorithm can be developed using minimum private data from users.
\section{Methodology - \texorpdfstring{$(P^{3}FitRec)$}{(P3FitRec)}}
\label{sec:method}

In this section, we first introduce an overview of our proposed framework ($P^{3}FitRec$), and then we introduce the details of each component of the proposed framework.

Our $(P^{3}FitRec)$ is composed of three-dimensional tensor data analysis and two deep learning models (MLP and multi-layer Bi-LSTM), which are used to predict the total workout distance, and the speed and heart rate sequences respectively. The main inputs of the framework are \textit{Target Caloric Expenditure, Sport Type, User ID}, and \textit{Workout Route ID}. Based on the \textit{User ID}, \textit{User Embedding} can be found by looking up the pre-trained user embedding tensor using the Tensor Decomposition method. Similarly, based on the \textit{Workout Route ID}, \textit{Workout Route Embedding} can be found by looking up the pre-trained workout route embedding tensor, as well as the \textit{Total Workout Route Distance, Altitude Sequence} and \textit{Distance Sequence} associated with the chosen route. In summary, the contextual input features consist of \textit{Target Caloric Expenditure, Sport Type, User Embedding, Workout Route Embedding}, and \textit{Total Workout Route Distance}. The sequential input features encompass \textit{Altitude Sequence} and \textit{Distance Sequence}. 

\subsection{Model Structure}

\begin{figure}[t]
\centering
\includegraphics[width=\textwidth]{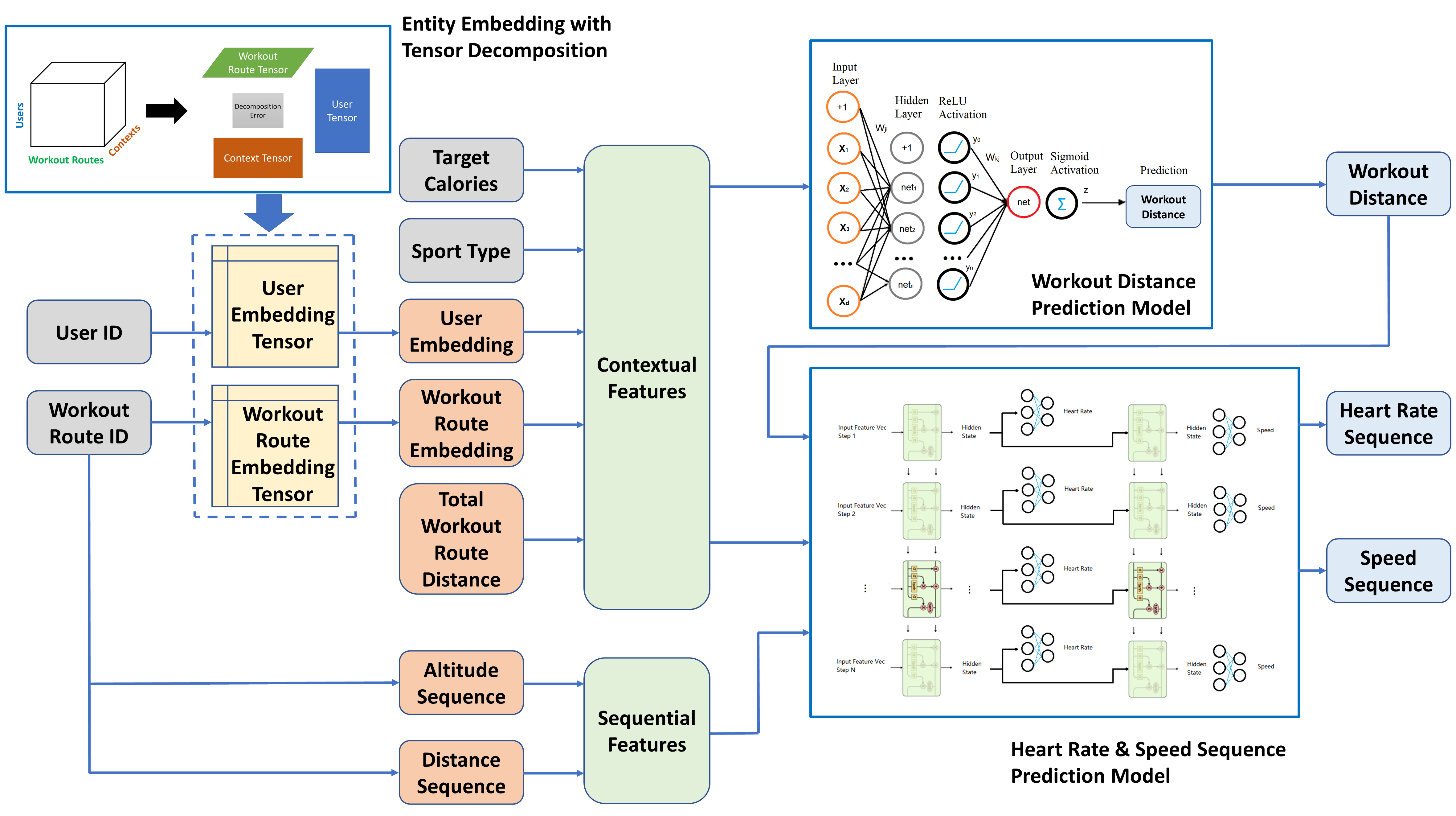}
\caption{Overview of our proposed framework $(P^{3}FitRec)$}
\label{fig:galaxy}
\end{figure}

Firstly, the MLP model takes the input of the contextual features to predict the required \textit{Total Workout Distance}. Subsequently, the predicted \textit{Total Workout Distance} is then concatenated with other contextual input features and sequential input features to form the input at each time step of the Multi-layer Bi-LSTM model. The Multi-layer Bi-LSTM model predicts the \textit{Heart Rate Sequence} through a fully connected layer at the output of the first LSTM layer and predicts the \textit{Speed Sequence} through another fully connected layer at the output of the second LSTM layer.

\subsection{Entity Embedding with Tensor Decomposition}

One of the main objectives of this study is to learn the latent characteristics of users from historical workout records without using their private data. We propose using a collaborative filtering method based on Tensor Decomposition to achieve this goal, which is a generalization of the conventional matrix decomposition method in higher-dimensional space.

For tensor analysis, a three-dimensional "user-workout route-contexts" tensor of size $N_\text{user} \times N_\text{route} \times N_\text{contexts}$ is constructed based on historical workout records. However, each workout record in the dataset contains a unique workout route because it has a unique sequence of altitudes, longitudes, and latitudes. To reduce the dimension of the three-dimensional tensor, we first cluster the workout routes into a smaller number of categories $N'_\text{route}$ where $N'_\text{route} < N_\text{route}$ to represent the workout routes in the three-dimensional tensor. In the third dimension, the contexts encompass several computed features like \textit{user’s gender, sport type, user’s workout frequency, user’s average workout duration, user’s average workout distance, user’s average workout speed, user’s average workout heart rate}, etc.

\begin{figure}[t]
\centering
\includegraphics[width=12cm]{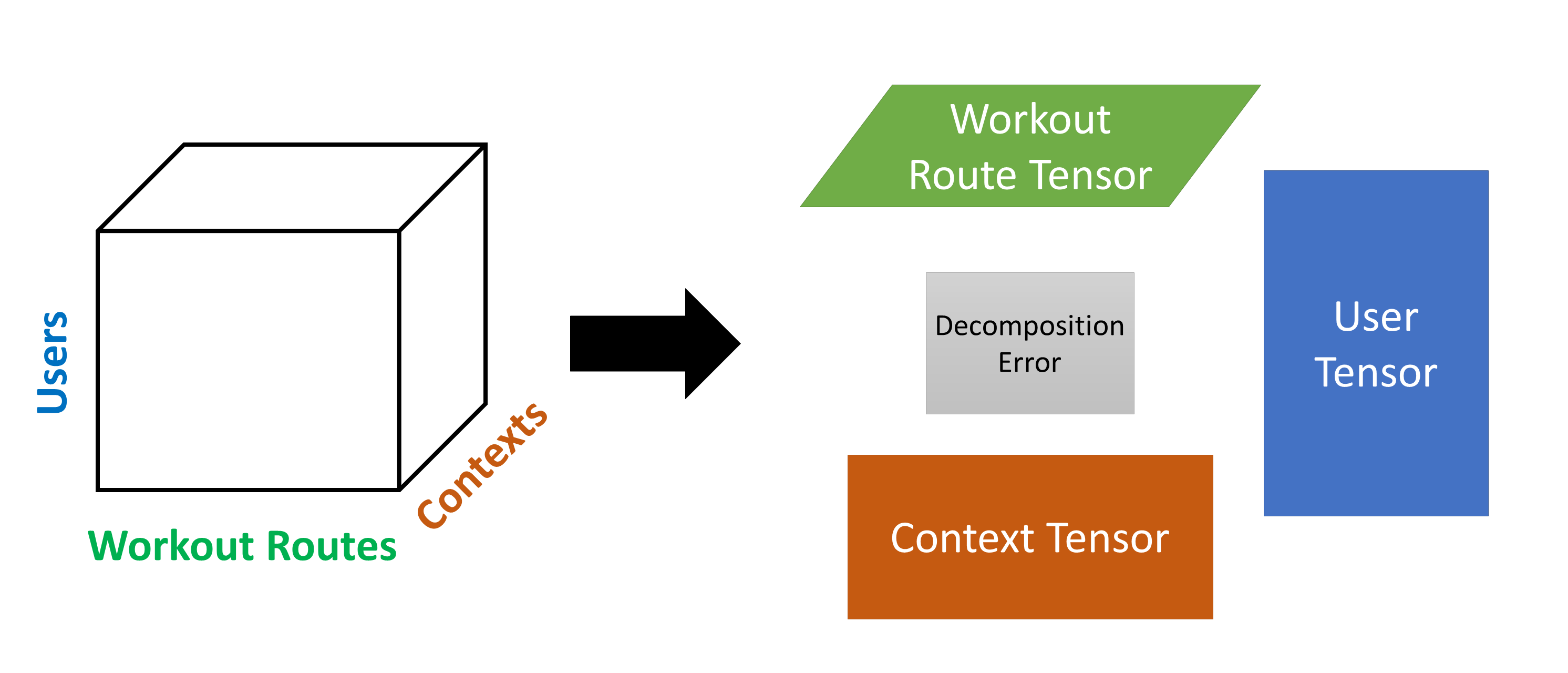}
\caption{"User-workout route-contexts" Tensor Decomposition}
\label{fig:Tensor_Decomposition}
\end{figure}

Several methods have been proposed in the literature for learning tensors known as tensor decomposition among which two typical approaches are mostly used in the literature known as CANDECOMP/ PARAFAC (CP) and Tucker decomposition \cite{kolda2009tensor}. This paper implements tensor decomposition using the CP approach since it has gained much popularity over tucker decomposition, and it is the most widely used algorithm due to its ease of interpretation.

Given a tensor $X \in \Re^{I \times J \times K} $, the main goal of  CP decomposition is to decrease the sum square error between the model and a given tensor $X$:
\begin{eqnarray}\label{cp}	
X  \approx \sum_{r=1}^R \lambda_r 	\ A_{r} \circ B_{r} \circ C_{r}  \equiv[ \lambda; A,B,C]
\end{eqnarray}
where "$\circ$" is a vector outer product. $R$ is the latent element, $A_{r}, B_{r}$ and $C_{r}$ are r-th columns of component matrices $A \in \Re^{I \times R}$, $B \in \Re^{J \times R} $and $ C \in \Re^{K \times R}$, and $\lambda$ is the weight used to normalize the columns of $A, B,$ and $ C $. 

In this sense, CP method decomposes $X$ into three  matrices $A$, $B$ and $C$ as shown in Fig. \ref{fig:Tensor_Decomposition}. Matrix $A$ represents the user mode, $B$ represents the workout route mode, and $C$ represents the context mode. This can be solved by minimizing the sum square error of

\begin{equation} \label{eq:als}
\min _{A, B, C}\left\|X-\sum_{r=1}^{R} \lambda_{r} A_{r} \circ B_{r} \circ C_{r}\right\|_{f}^{2}
\end{equation}

At first, the function given in Equation \ref{eq:als} seems to be a non-convex problem, because its goal is to optimize the sum of squares of three matrices. However, by fixing two factor matrices and solving only the third one, this problem can be simplified to a linear least squares problem. Following this approach, the ALS technique can be employed here, which solves every component matrix repeatedly by locking all other components until it converges \cite{anaissi2018regularized}. 







\begin{equation} \label{eq:cp decomposition 2}
X=A L(C \odot B)^{T}+E
\end{equation}

Assume we have completed CP Decomposition with a selected rank and learned matrices $A$, $B$, and $C$. Then we fit the full Tucker3 model to the data using the CP Decomposition matrices $A$, $B$, and $C$ by minimizing

\begin{equation} \label{eq:cp decomposition 3}
\sigma(\mathbf{G})=\left\|\mathbf{X}-\mathbf{A} \mathbf{G}(\mathbf{C} \otimes \mathbf{B})^{\mathrm{T}}\right\|_{\mathrm{F}}^{2}
\end{equation}

where $\otimes$ denotes Kronecker product.

The optimal $\mathbf{G}$ in equation \ref{eq:cp decomposition 3} can be determined as

\begin{equation} \label{eq:cp decomposition 4}
\operatorname{vec} \mathbf{G}=(\mathbf{C} \otimes \mathbf{B} \otimes \mathbf{A})^{+} \operatorname{vec} \mathbf{X}
\end{equation}

when the residual decomposition error $\sigma(\mathbf{G})$ is minimized. 

The underlying idea is to find the similarity between $\mathcal{L}$ and $\mathcal{G}$ where $\mathcal{L}$ is a super diagonal core tensor that all its super diagonal values are 1 and all its off-superdiagonal values are 0. To compare the similarity, we can have a look at the distribution of the elements in the superdiagonal and off-super diagonal of $\mathcal{G}$. If the superdiagonal elements of $\mathcal{G}$ are all close to the corresponding elements of $\mathcal{L}$, which is 1, and the off-superdiagonal elements of $\mathcal{G}$ are all close to the corresponding elements of $\mathcal{L}$, which is 0, then we say the CP Decomposition result is appropriate. Formally, the similarity between the two tensors, or core consistency can be written as

\begin{equation} \label{eq:cp decomposition 5}
c c=100\left(1-\frac{\sum_{l=1}^{R} \sum_{m=1}^{R} \sum_{n=1}^{R}\left(g_{l m n}-\lambda_{l m n}\right)^{2}}{R}\right)
\end{equation}

where the closer the cc score is to 100 the better.

\subsection{Workout Distance Prediction}

Our distance prediction model is based on MLP architecture, which refers to a fully connected neural network containing one or more hidden layers. As MLP is the basic form of neural network that can be used to solve classification and regression tasks, we adopt the MLP architecture to build a model that predicts workout distance using the contextual input features $X_\text{context} = [x_\text{user\_embed}, x_\text{route\_embed}, x_\text{context\_1}, x_\text{context\_2}, \cdots, x_\text{context\_n}]$. 

\begin{figure}[t]
\centering
\includegraphics[width=12cm]{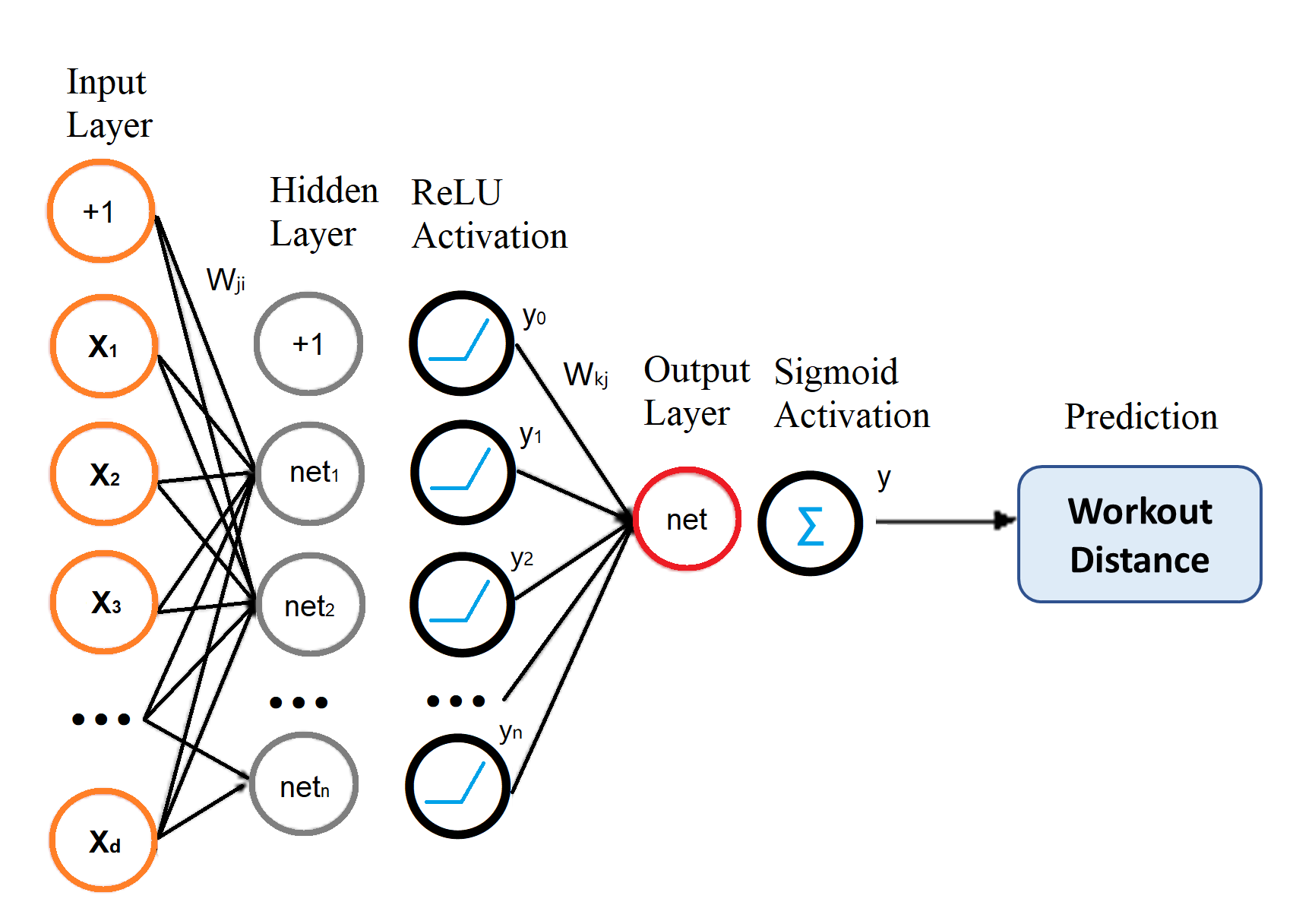}
\caption{Our Workout Distance Prediction Model with MLP}
\label{fig:MLP_Model_Structure}
\end{figure}

For the MLP model shown in Fig. \ref{fig:MLP_Model_Structure}, we feed the contextual input features $X_\text{context}$ into the network, and the output at the hidden layer $\text{net}_j$ can be calculated by by\begin{equation} \label{eq:mlp 1}
\text{net}_j=W_j X_\text{context} + B_j
\end{equation}
The output of hidden layer after activation is $y_j = \text{ReLU}(\text{net}_j)$ where
\begin{equation} \label{eq:mlp 2}
\text{ReLU}(x) = \text{max}(0, x)
\end{equation}
Similarly, we can compute the output at the output layer by
\begin{equation} \label{eq:mlp 3}
\text { net }_k=W_k y_{k-1} + B_k
\end{equation}
And finally the model predicts a normalized workout distance by 
\begin{equation} \label{eq:mlp 4}
y_k = \text{Sigmoid}(\text { net })
\end{equation}
where
\begin{equation} \label{eq:mlp 5}
\text{Sigmoid}(x) = \frac{1}{1+e^{-x}}
\end{equation}

\subsection{Speed and Heart Rate Sequence Prediction}

We propose a Multi-layer Bi-LSTM Model to predict speed and heart rate sequences due to inputs and outputs being in sequential format. LSTM-based models are an extension for RNNs that implement memory states to store information and gate mechanisms to control information flow for the purpose of alleviating vanishing gradient problems \cite{9005997}. More specifically, they use cell state and hidden state to carry information. A forget gate is used to control the preservation and removal of information passed from the last time step. Meanwhile, an input gate is used to decide what new information we’re going to store, while an output gate is used to specify what information contributes to the output at the current time step.

Firstly, the output through the forget gate $f_{t}$ is computed as:
\begin{equation} \label{eq:lstm 1}
f_{t}=\sigma\left(W_{f}\left[h_{t-1}, x_{t}\right]+b_{f}\right)
\end{equation} 

where
\begin{list}{$\circ$}{} 
\item $h_t$, $h_{t-1}$ are the hidden states of the LSTM at time step $t$ and $t-1$ respectively
\item $x_t$ is the input at time step $t$
\item $\sigma$ denotes sigmoid activation function
\end{list}
Then the output through the input gate $\tilde{C}_t$ is computed as:
\begin{equation}\label{eq:lstm 2}
    \begin{cases} i_{t}=\sigma\left(W_{i}\left[h_{t-1}, x_{t}\right]+b_{i}\right)\\ \tilde{C}_t=\tanh\left(W_{C}\left[h_{t-1}, x_{t}\right]+b_{C}\right)
    \end{cases}       
\end{equation}
where $\tanh$ denotes hyperbolic tangent activation function.
Meanwhile, the cell state from the previous time step $C_{t-1}$ is updated by:
\begin{equation} \label{eq:lstm 3}
C_{t}=f_t * C_{t-1} + i_t * \tilde{C}_t
\end{equation}

\begin{figure}[t]
\centering
\includegraphics[width=12cm]{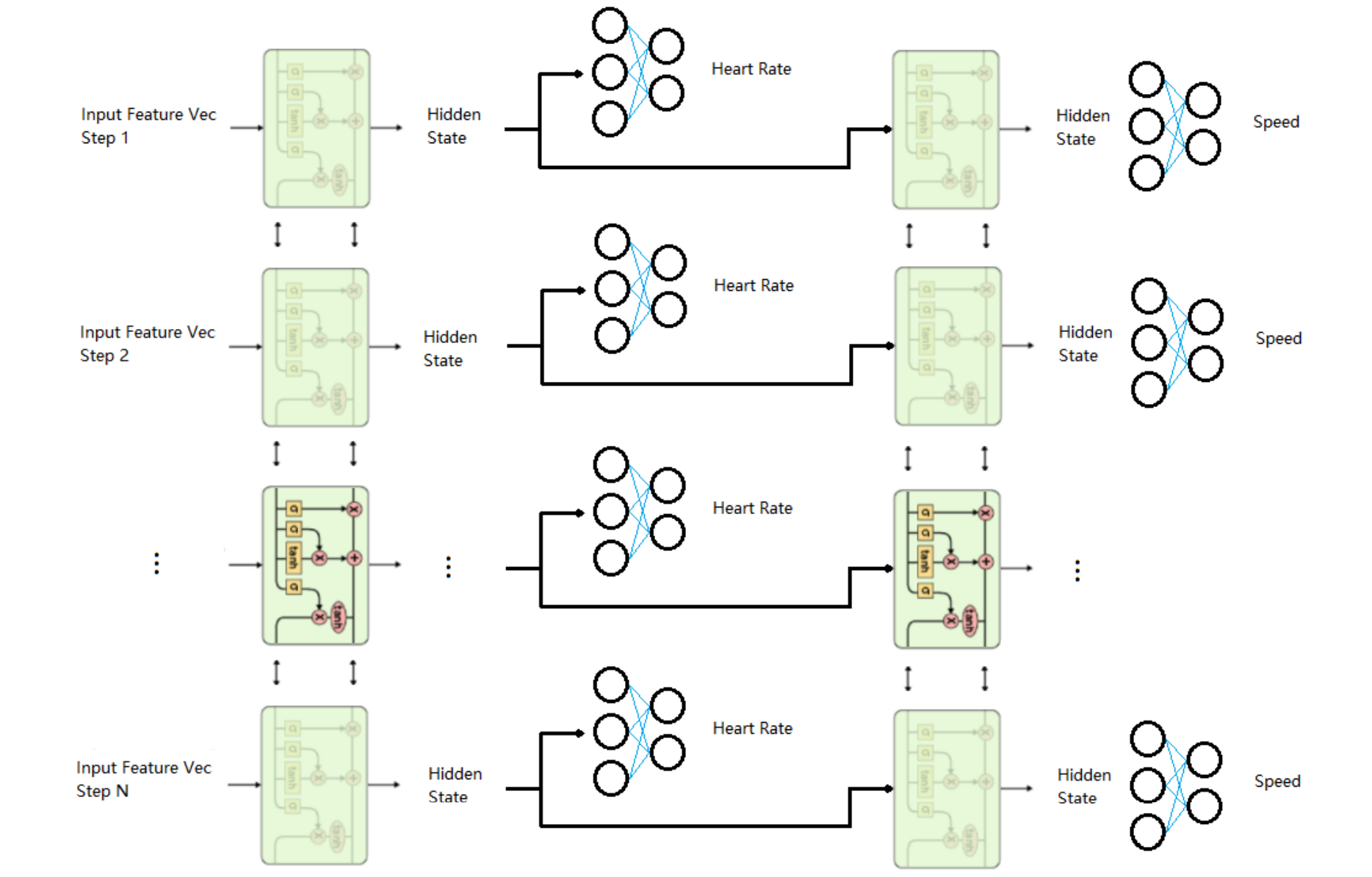}
\caption{Speed and HeartRate Model}
\label{fig:Speed_HeartRate_Model}
\end{figure}

Lastly, the output at the current time step $h_{t}$ is computed as:

\begin{equation}\label{eq:lstm 4}
    \begin{cases} o_{t}=\sigma\left(W_{o}\left[h_{t-1}, x_{t}\right]+b_{o}\right)\\ 
    h_{t}=o_t + \tanh (C_t)
    \end{cases}       
\end{equation}
\newline
Fig. \ref{fig:Speed_HeartRate_Model} shows the detailed structure of the 2-layer Bi-LSTM model. At step $t$, with input $X_t = [X_\text{context}, x_{\text{altitude}\_t}, x_{\text{distance}\_t}]$, heart rate $y_{\text{heart\_rate}\_t}$ is predicted through the hidden state of the first Bi-LSTM layer and speed $y_{\text{speed}\_t}$ is predicted through the hidden state of the second Bi-LSTM layer. We propose predicting the heart rate and speed at two different stages instead of predicting them both at the second stage or just using a single LSTM stage.

When training the models, we find that the predicted heart rate sequence is always positively correlated with the input calories. However, in some cases, the predicted speed might be negatively correlated to the calorie input during inference, which is counter-intuitive. Therefore, we design the model structure by predicting the heart rate at the first layer and then using its hidden states as the input to the second LSTM layer to predict the speed. This approach reinforces the model to learn the correct correlation between speed and input calories to alleviate the counter-intuitive problem.

Formally, if we denote the output of LSTM as $h^{(i)}$ where $i$ refers to $i^{th}$ layer, the predicted heart rate and speed are computed as:

\begin{equation}\label{eq:lstm 5}
    y^{(i)} = \text{SELU}(W^{(i)} h^{(i)} + b^{(i)})
\end{equation}

where SELU denotes scaled exponential linear unit activation function. Inspired by \cite{NIPS2017_5d44ee6f} and \cite{10.1145/3308558.3313643}, our sequence model adopts SELU as the activation function, because it induces self-normalizing properties like variance stabilization, which can solve vanishing and exploding gradient problems.


Furthermore, the use of the bidirectional mechanism helps the model more effectively learn the underlying context by traversing the sequential input features twice, i.e., both forward and backward \cite{9005997}. For instance, understanding whether the workout route is going uphill or downhill facilitates a better prediction of speed and heart rate.
\section{Experiments and Results}
\label{sec:evaluation}

\subsection{Dataset}
\label{sec:dataset}
We use the open-source \textbf{Endomondo} dataset \cite{10.1145/3308558.3313643} of 167,373 workout records from 956 users. Each workout record consists of both sequential data (i.e., \textit{heart rate, speed, time elapsed, distance, altitude, latitude, longitude}) and contextual data (i.e., \textit{gender, sport, userID, URL}). For each workout record, the sequential data contains 500 data points, with sampling intervals ranging from seconds to minutes. In addition, the total \textit{caloric expenditure} of each workout record can be queried on endomondo.com, if the user is not deactivated.

We have conducted data cleaning prior to the experiments. For instance, the original dataset has a high rate of abnormal measurements such as running speeds exceeding 50 km/h or abnormal average altitudes over 8000 meters. Moreover, 97$\%$ of the workout records belong to the types of running, cycling, and mountain biking, while there are fewer than 50 records for most other sport types in the dataset. Hence, we decide to only focus on running, cycling, and mountain biking in our study to ensure there is enough data for each sport type. About 65,500 workout records are retained after cleaning up and deleting the records that no longer provide \textit{caloric expenditure} due to users' discontinuation. 

Data augmentation is done to facilitate the modeling of workout distance according to different input variables. For instance, we expect our model outputs a shorter distance given a smaller caloric input. Specifically, we extend a workout route by appending a small random sub-sequence extracted from the start of the sequence to the end of the sequence if the user has returned to the starting location at the end of the exercise. Given a sequence $X = [x_0, x_1, x_2, x_3, \cdots, x_n, x_0]$ that returns to the starting location $x_0$, the extended sequence can be represented as $X' = [x_0, x_1, x_2, x_3, \cdots, x_n, x_0, x_1, \cdots, x_t]$ where $t < n$. For the workout distance prediction model, the original workout distance $L_X$ is used as the ground truth while the caloric expenditure, the extended workout distance $L_{X'}$, the entity embeddings, and the other contextual features are model inputs. 
The resulted dataset after the our pre-processing will be made available from the project repository for further experimentation and reproducibility.  \protect\footnote{\url{https://github.com/BasemSuleiman/Personalized\_Intelligent\_Fitness\_Recommender}.}

\subsection{Training Procedure}
\label{sec:training_procedure}

We first perform CP tensor decomposition to obtain user embeddings and workout route embeddings. Core consistency scores of ranks 2 to 20 are computed to choose the most appropriate rank of the decomposed tensors. Then we pick the entity embeddings of high scores to train the workout distance prediction model and the speed \& heart rate sequences prediction model respectively. The workout distance prediction model is trained using Adam \cite{adam} with weight decay of $1e^{-7}$ and the speed \& heart rate prediction model is trained using Adagrad \cite{adagrad}. For both models, we perform hyperparameter tuning with random search and/or grid search on the validation set. The best parameters are shown in table \ref{table:parameter_settings}.

\begin{table}[t]
\centering
\begin{tabular}{llll} \toprule

& 
\makecell[l]{\textbf{Workout Distance} \\ \textbf{Model}} &
&
\makecell[l]{\textbf{Speed \& HeartRate} \\ \textbf{Model}} \\ \midrule

2 Layer MLP &
\begin{tabular}{@{}l@{}} Learning Rate: $1e^{-3}$ \\ Hidden Dimension: $64$ \\ Dropout: $0.2$ 
\end{tabular} &
1 Layer LSTM &
\begin{tabular}{@{}l@{}} Learning Rate: $5e^{-3}$ \\ Hidden Dimension: $64$ \\ Dropout: $0.2$ 
\end{tabular} \\ \midrule

3 Layer MLP & 
\begin{tabular}{@{}l@{}} Learning Rate: $1e^{-3}$ \\ Hidden Dimension 1: $64$ \\
Hidden Dimension 2: $64$ \\ Dropout: $0.2$ 
\end{tabular} &
2 Layer (Bi) LSTM & 
\begin{tabular}{@{}l@{}} Learning Rate: $5e^{-3}$ \\ Hidden Dimension 1: $128$ \\
Hidden Dimension 2: $64$ \\ Dropout: $0.2$ 
\end{tabular} \\ \bottomrule
 
\end{tabular}
\caption{Parameter settings}
\label{table:parameter_settings}
\vspace{-8mm}
\end{table}

\subsection{Evaluation Metrics}
\label{sec:evaluation_metrics}

We report the results of our experiments through Root Mean Squared Error (RMSE) and Mean Absolute Error (MAE) for the workout distance prediction task and speed \& heart rate sequences prediction task respectively.

\begin{equation}
\text{RMSE}=\sqrt{\frac{1}{N_\text{test}} \sum_{y \in \mathcal{T}_{\text {test }}}\left(y_{t}-\hat{y}_{t}\right)^{2}}
\end{equation}

\begin{equation}
\text{MAE}=\frac{1}{N_\text{test}} \sum_{y \in \mathcal{T}_{\text {test }}} \frac{1}{L} \sum_{t=1}^{L}\left|y_{t}-\hat{y}_{t}\right|
\end{equation}

where $N_\text{test}$ is the number of  workout records in the test set $\mathcal{T}_{\text {test }}$ and $L$ is the number of time steps in each workout record. 

\subsection{Entity Embedding with Tensor Decomposition}
\label{sec:tensor_decomposition_eval}

As discussed in section \ref{sec:method}, we adopt the CP tensor decomposition method to generate entity embeddings and perform core consistency diagnostic to find the best rank of the decomposed tensors. Using formula \ref{eq:cp decomposition 5} to compute core consistency, the closer the value is to 100, the more appropriate the rank is. Generally speaking, with the increase of the number of ranks, the core consistency score tends to decrease monotonically due to the increase of decomposition noise and other non-trilinear variations \cite{080bc1d7971d4328add2c543579ec1f4}.

\begin{figure}[t]
\centering
\includegraphics[width=12cm]{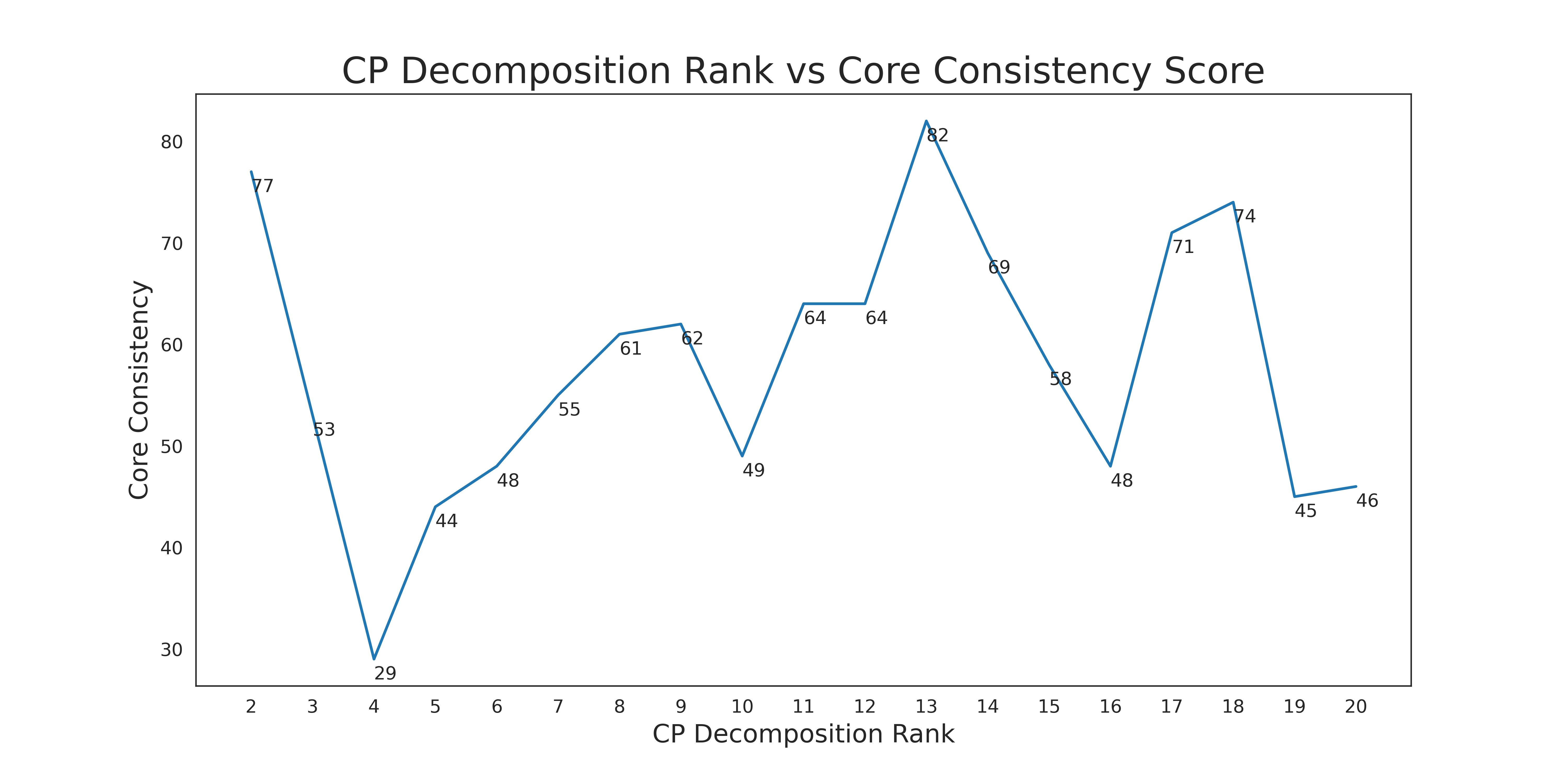}
\caption{CP Decomposition rank vs Core Consistency}
\label{fig:CP Decomposition rank vs Core Consistency}
\end{figure}

As Figure \ref{fig:CP Decomposition rank vs Core Consistency} illustrates, the core consistency score is relatively high starting at CP Decomposition rank of 2. With the increase of CP Decomposition ranks, the score fluctuates and peaks at the rank of 13, after which a downward trend can be observed. 

Core consistency diagnostic provides an intrinsic evaluation of the CP Decomposition rank. However, it does not necessarily guarantee that the rank selected with this method will perform best at the final regression task. Therefore, we select the decomposed entity embedding tensors of rank 2, 11, and 13, respectively, and use the final models to further evaluate them. The rank 13 tensors have the highest core consistency score, and they are relatively high in dimension, which may increase challenges to the convergence of the final models. In contrast, the rank 2 tensors have a relatively high core consistency score and low dimension. We also pick rank 11 to balance the core consistency score and complexity.

Inspired by word2vec \cite{word2vec}, we evaluate the trained entity embeddings with cosine similarity: $\operatorname{similarity}(A, B)=\frac{A \cdot B}{\|A\| \times\|B\|}=\frac{\sum_{i=1}^{n} A_{i} \times B_{i}}{\sqrt{\sum_{i=1}^{n} A_{i}^{2}} \times \sqrt{\sum_{i=1}^{n} B_{i}^{2}}}$. Taking user embeddings of size 13 as an example, the cosine similarity between randomly picked user A (192 running records) and user B (129 running records) is 0.78660, while that between user A and user C (132 mountain biking and 1 biking records) is -0.00489. Similarly, the cosine similarity between user D (104 biking and 70 running records) and user A is 0.57067, and that between user D and user C is 0.18126. The findings suggest that sport type has a large impact on the trained user embeddings, that is, users playing similar sports may have closer embeddings. 

\begin{figure}[ht]
\centering
\includegraphics[width=\textwidth]{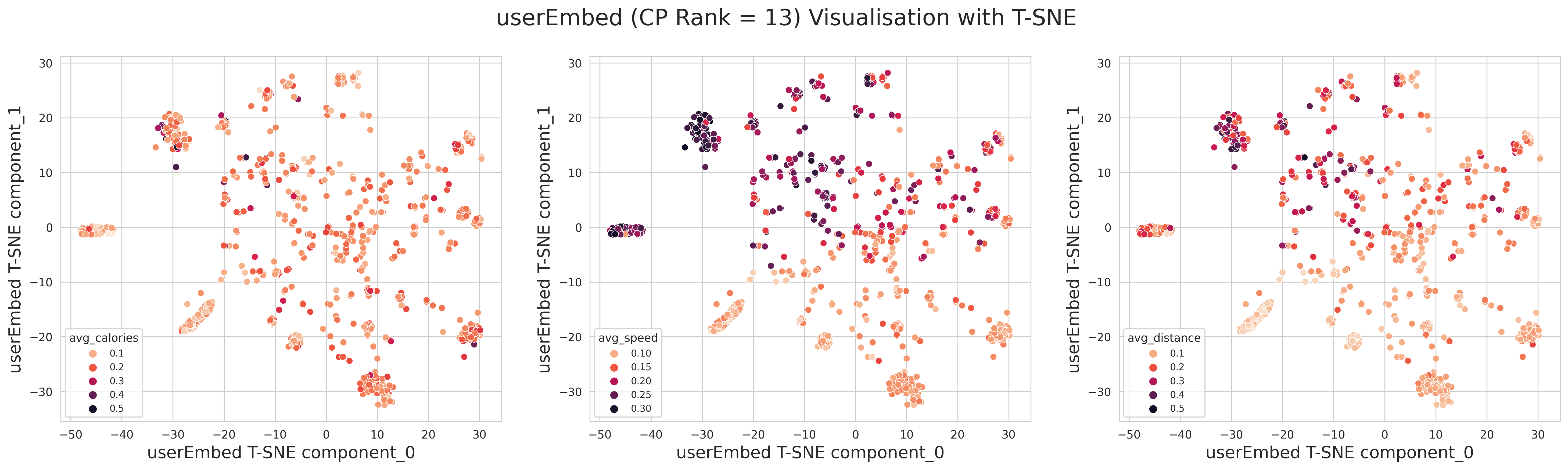}
\caption{Rank 13 User Embedding T-SNE Plot}
\label{fig:user_embed_example}
\end{figure}

\begin{figure}[ht]
\centering
\includegraphics[width=\textwidth]{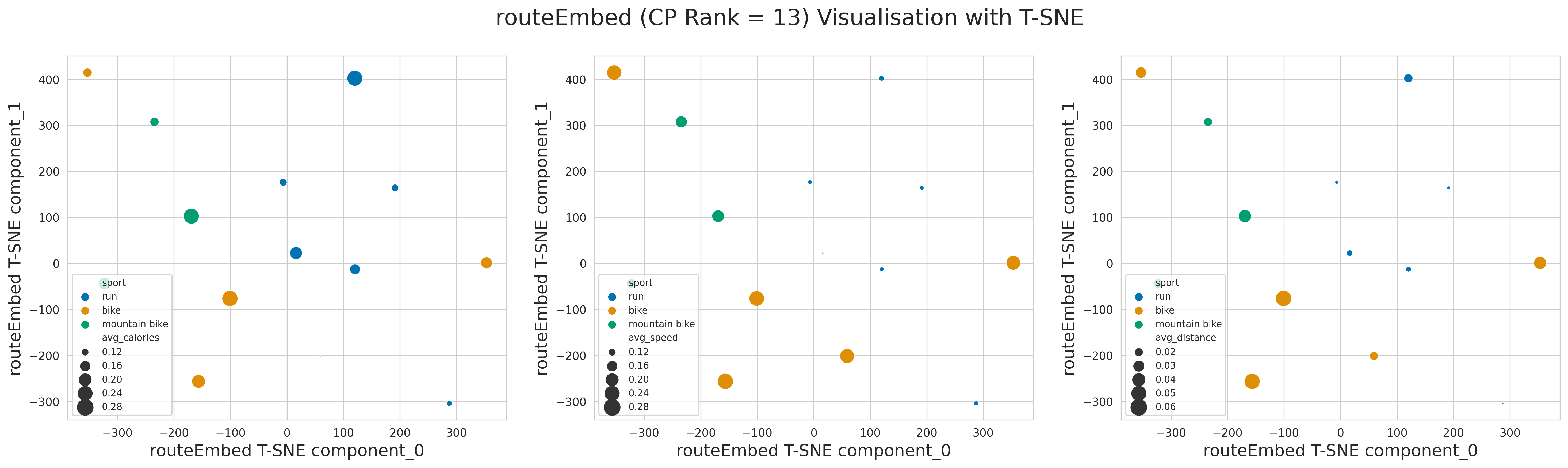}
\caption{Rank 13 Workout Route Embedding T-SNE Plot}
\label{fig:route_embed_example}
\end{figure}

In addition, we use T-SNE to plot the entity embeddings on a 2D plane, as shown in  Fig. \ref{fig:user_embed_example} and Fig. \ref{fig:route_embed_example}. To compare the results intuitively, 3 user embedding scatterplots are drawn, and color-coded according to the average workout calories, average workout speed, and average workout distance respectively. Similarly, we plot 3 figures for the workout route embeddings, and color code them according to the type of sport. The size of each data point in the figures is proportional to the magnitudes of the average workout calories, average workout speed, and average workout distance respectively. As for the user embeddings, we observe that the users with similar preferences of average workout speeds and average workout distances tend to be clustered together. Although patterns are not as obvious for the workout route embeddings, possibly due to a lack of data points, the points of the same sports seem to be clustered together for running and mountain biking. More T-SNE plots of the entity embeddings of other ranks can be found in APPENDICES: \ref{sec:appendices}.

\subsection{Workout Distance Prediction Model}

For the experiment of the distance prediction sub-model, we ran ablations on training MLP models with different hidden layers and different entity embedding dimensions as input features. The former is to heuristically seek the optimal number of hidden layers for best model performance, while the latter is an extrinsic evaluation of the pre-trained entity embeddings on solving one of our final tasks. 

\begin{table}[t]
\centering
\begin{tabular}{lc} \toprule
 
\textbf{Models}&\textbf{Distance RMSE (KM)} \\ \midrule

2 Layer MLP &  \\

$\;$ + Embedding Size 2 & 0.1423 \\
  
$\;$ + Embedding Size 11 & 0.1394 \\
  
$\;$ + Embedding Size 13 & \textbf{0.1387} \\

3 Layer MLP &  \\

$\;$ + Embedding Size 2 & 0.1422 \\
  
$\;$ + Embedding Size 11 & 0.1420 \\
  
$\;$ + Embedding Size 13 & 0.1433 \\ 
\bottomrule
 
\end{tabular}
\caption{Ablation study of MLP layers and entity embedding dimensions}
\label{table:ablation_study_mlp}
\vspace{-4mm}
\end{table}

As shown in Table \ref{table:ablation_study_mlp}, the model that produces the best performance is 2 Layer MLP (1 hidden layer) with entity embedding of size 13. Since the performances of the 3 Layer MLP models (2 hidden layers) do not exceed those of the 2 Layer MLP models, we choose not to further experiment with more hidden layers. The task of predicting workout distance produces a single scalar, which is relatively simple. Hence, a 2 Layer MLP model is adequate. In contrast, the increase of hidden layers is likely to lead to over-fitting.

Moreover, the result of the extrinsic evaluation of entity embeddings is consistent with the intrinsic evaluation result in section \ref{sec:tensor_decomposition_eval}. Regardless of the number of MLP layers, the best model is achieved with entity embedding of size 13, followed by that of size 2, and then that of size 11. In the intrinsic evaluation of entity embeddings, we found that entity embedding of size 13 has the highest core consistency score, followed by that of size 2, and then that of size 11. This finding validates the hypothesis of core consistency, that is, the higher the core consistency value, the more appropriate the CP decomposition result is.

Although we cannot compare the result of this model with FitRec \cite{10.1145/3308558.3313643}, as it does not have the task of predicting workout distance, our best model achieves an RMSE of 0.1387 (km), which is reasonably good for predicting workout distance in practice.

\subsection{Speed \& Heart Rate Prediction Model}

For the speed \& heart rate prediction sub-model, we first ran ablations on different entity embedding inputs based on a 1 Layer LSTM network structure, which is an extrinsic evaluation of the pre-trained entity embeddings. From Table \ref{table:ablation_study_lstm}, we observe that MAE drops as embedding size increases. The 1 Layer LSTM model with an entity embedding size of 13 is found to yield the best performance, which decreases the MAE of the baseline (1 Layer LSTM without entity embedding) by 5.8$\%$ and 12.5$\%$ on speed and heart rate respectively. The result of this experiment is consistent with the core consistency diagnostic result in section \ref{sec:tensor_decomposition_eval} that the embedding size of 13 is most appropriate. 

\begin{table}[t]
\centering
\begin{tabular}{lcc} \toprule
 
\makecell[l]{\textbf{Speed \& HeartRate} \\ \textbf{Model}} & \makecell[c]{\textbf{Speed MAE} \\ \textbf{(KMPH)}} & \makecell[c]{\textbf{Heart Rate MAE} \\ \textbf{(BPM)}} \\ \midrule
  
1 Layer LSTM &  &  \\

$\;$ + No Embedding & 2.92 & 13.01 \\
  
$\;$ + Embedding Size 2 & 2.90 & 13.01  \\
 
$\;$ + Embedding Size 11 & 2.80 & 11.57  \\
 
$\;$ + Embedding Size 13 & \textbf{2.75} & \textbf{11.38}\\ \toprule

Embedding Size 13 &  &  \\

$\;$ + 2 Layer Stacked LSTM & 2.51 & 11.376 \\

$\;$ + 2 Layer Stacked Bi-LSTM & \textbf{2.4} & \textbf{11.304} \\

$\;$ + 2 Layer Stacked Bi-LSTM + Attention & 3.2 & 13.92  \\

\bottomrule
  
\end{tabular}
\caption{Ablation study of LSTM structure and entity embedding dimensions}
\label{table:ablation_study_lstm}
\vspace{-8mm}
\end{table}

Then we ran ablations on different LSTM structures with entity embedding size set to 13. As shown in Table \ref{table:ablation_study_lstm}, model performances are improved by applying another layer of LSTM and the bidirectional LSTM structure. The best model is the 2 Layer Stacked Bi-LSTM with embedding size 13, whose MAEs on speed and heart rate are 2.4 km/h and 11.3 beats/minute respectively. Compared with the baseline model (1 Layer LSTM without entity embedding), its MAEs are 17.8\% and 13.1\% less for speed and heart rate predictions respectively. However, the addition of the attention mechanism does not improve model performance. It is likely that speed and heart rate mostly depend on neighboring steps rather than distant steps in the sequence. In this case, the attention mechanism has little help since it primarily addresses information loss problems for long sequences \cite{attention}.

\subsection{Comparison with FitRec}

To further evaluate our result, we compare our $(P^{3}FitRec)$ with  FitRec \cite{10.1145/3308558.3313643}. As shown in Table \ref{table:comparison_with_FitRec}, our $(P^{3}FitRec)$ and FitRec use the same dataset, and both aim to provide fitness recommendations by predicting speed and heart rate sequences. The comparison is conducted in the following aspects. First, in pre-processing, FitRec keeps all sports while we only keep running, cycling, and mountain biking, which account for 97\% of the total workouts, considering each sport needs sufficient samples to train a decent deep learning model. In addition, for each workout, FitRec only extracts the first 450 out of the 500 timestamps to reduce noise in the dataset, while we recommend keeping the entire 500 timestamps so that the model is more practical in real life. Furthermore, both studies share similar input features, including historical user information, route information, sport type, and gender. However, FitRec uses the time sequence and distance sequence as their inputs to predict speed sequence, while we remove the time sequence to avoid data leakage ($\text{speed} = \frac{\text{distance}}{\text{time}}$), which adds a significant challenge to the regression task. Considering the practical application of the model, we also add target caloric consumption to be one of the input features.

Moreover, both systems derive user embeddings to extract the latent characteristics of the users from historical workout data. However, FitRec only uses a user's latest exercise record to derive user embeddings through an LSTM network. In contrast, we train user embeddings and workout route embeddings through tensor decomposition with all historical workout records. Using more records enables us to realize more abundant information from the entities.

\begin{table}[ht]
\centering
\begin{tabular}{lll} 
\toprule

& 
\textbf{$(P^{3}FitRec)$} & \textbf{FitRec} \\ \midrule

\textbf{Pre-processing} & 
\parbox{.35\textwidth}{
\begin{itemize}
  \item 3 sport types (Account for 97\% of data)
  \item Entire sequence of each workout
\end{itemize}
} 

&
\parbox{.35\textwidth}{
\begin{itemize}
  \item All sport types
  \item First 450 of 500 timestamps for each workout
\end{itemize}
} \\ \midrule

\textbf{Input and Output} &

\begin{tabular}[t]{@{}l@{}}
Input: \\

\parbox{.35\textwidth}{
\begin{itemize}
  \item Sport type
  \item Gender
  \item Altitude sequence
  \item Distance sequence
  \item \textbf{Calories}
  \item User embedding \textbf{(Tensor decomposition)}
  \item Route embedding \textbf{(Tensor decomposition)}
\end{itemize}
} \\

Output: \\

\parbox{.35\textwidth}{
\begin{itemize}
  \item Speed sequence
  \item Heat rate sequence
\end{itemize}
}
\end{tabular}

& 
\begin{tabular}[t]{@{}l@{}}
Input: \\

\parbox{.35\textwidth}{
\begin{itemize}
  \item Sport type
  \item Gender
  \item Altitude sequence
  \item Distance sequence
  \item \textbf{Time sequence}
  \item User embedding \textbf{(LSTM)}
\end{itemize}
} \\

Output: \\

\parbox{.35\textwidth}{
\begin{itemize}
  \item Speed sequence
  \item Heat rate sequence
\end{itemize}
}

\end{tabular}

\\ \midrule

\textbf{Embedding method} &

\parbox{.35\textwidth}{
\begin{itemize}
  \item Trained on all historical records
  \item Tensor decomposition
\end{itemize}
} &

\parbox{.35\textwidth}{
\begin{itemize}
  \item Trained on the user's most recent record
  \item LSTM
\end{itemize}
} \\ \midrule

\textbf{Best Model} &

\begin{tabular}[c]{@{}l@{}}
2-layer stacked Bi-LSTM \\ (Speed \& Heart Rate \\ 
by single model)
\end{tabular} & 

\begin{tabular}[c]{@{}l@{}}
2-layer stacked LSTM \\ (Speed \& Heart Rate \\ by 2 models)
\end{tabular} \\ \midrule

\textbf{Results} &

\begin{tabular}[c]{@{}l@{}}
Speed MAE: 2.4 KM/H; \\ Heart Rate MAE: \\ 11.304 BPM
\end{tabular} &

\begin{tabular}[c]{@{}l@{}}
Speed MAE: 2.384 KM/H; \\ Heart Rate MAE: \\ 12.847 BPM
\end{tabular} \\ \bottomrule
\end{tabular}

\caption{Comparison between $(P^{3}FitRec)$ and FitRec}
\label{table:comparison_with_FitRec}
\vspace{-4mm}
\end{table}

Next, our speed and heart rate prediction model has a bidirectional LSTM structure while FitRec implements a unidirectional LSTM.

Finally, FitRec and our model have achieved similar performance in predicting speed and heart rate. More specifically, FitRec performs a little better in speed prediction, while our result in heart rate prediction is better. This does not mean our model is inferior in predicting speed. FitRec uses distance and elapsed time as inputs to predict speed, which might lead to data leakage to a certain extent. The removal of the elapsed time feature in our implementation adds challenges to the speed prediction task.

Moreover, we also add target caloric expenditure as an input feature to our model to influence the magnitudes of the predicted speed sequence, which potentially further elevates the difficulty of the task. Despite these challenges, we still achieve a comparative result in speed prediction and a superior result in heart rate prediction. Furthermore, the implementation of tensor decomposition provides a more flexible and lighter method for extracting user and route information. The overall better result can be due to many factors, such as a better strategy of training entity embeddings, different model design, and less variance of sport types in the dataset, etc. 

The proposed models, datasets and experimental data will be made available at the project repository for further extension and reproducibility studies. \protect\footnote{\url{https://github.com/BasemSuleiman/Personalized\_Intelligent\_Fitness\_Recommender}.}.
\section{Discussion}
\label{sec:discussion}
\begin{figure}[t]
\centering
\includegraphics[width=12cm]{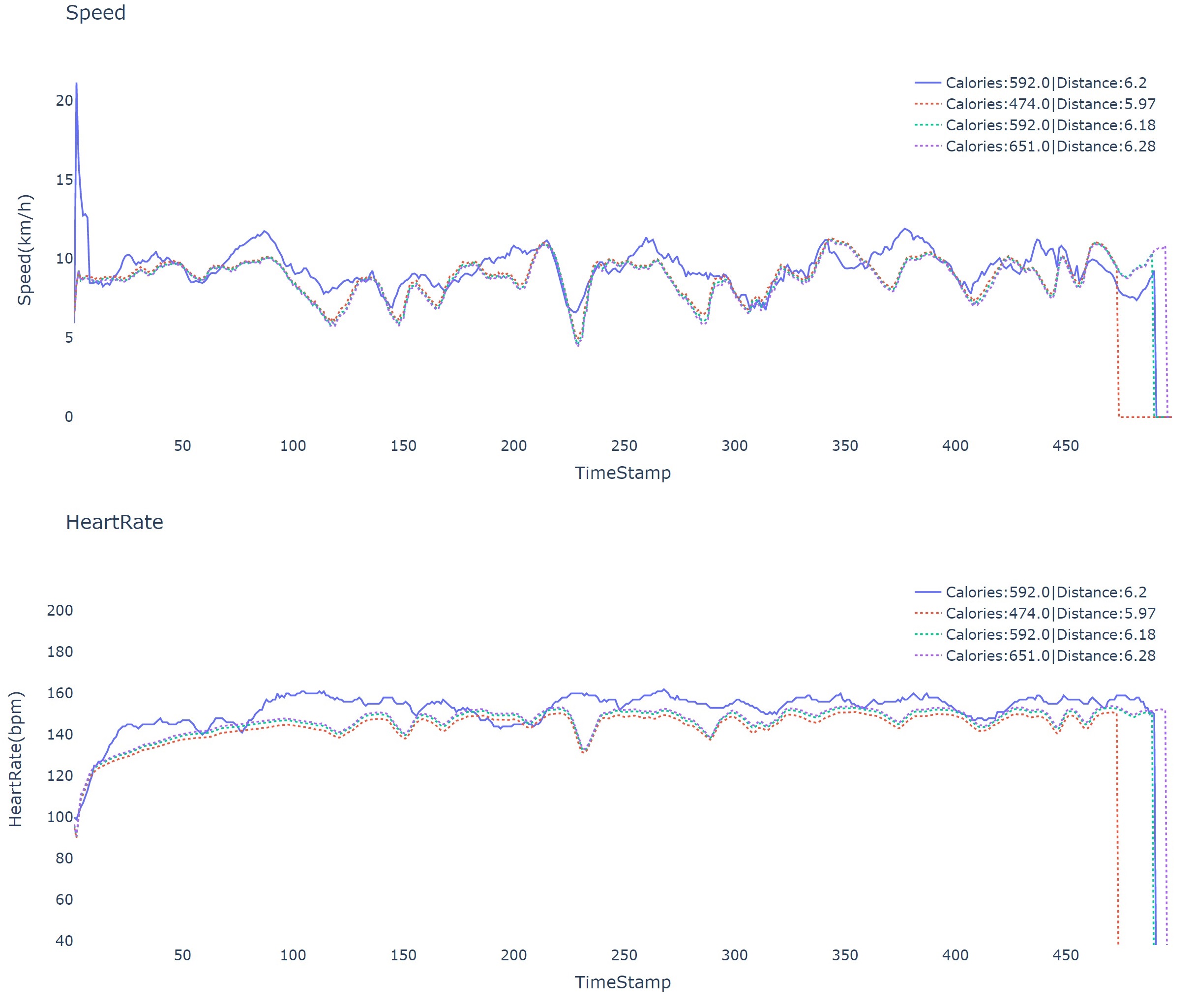}
\caption{Speed and Heart Rate Profile}
\label{fig:speed_hr_profile}
\end{figure}

Our $(P^{3}FitRec)$ can provide personalized fitness recommendations by applying the models mentioned in section \ref{sec:method}. The system allows users to specify the type of sport (run, bike, or mountain bike) and select a specific route, and then input target caloric expenditure. According to these inputs, our $(P^{3}FitRec)$ will provide various recommendations in the aspects of distance as well as speed and heart rate at each time step. 
Table \ref{table:workout_profile_table} shows a typical running workout from the dataset, whose original record has 6.2 km of distance, 592 kcal of caloric consumption, and average speed and heart rate of 8.8 km/h and 149 beats per minute respectively. Given the same caloric input, our $(P^{3}FitRec)$ predicts similar distance, speed, and heart rate. Furthermore, for higher target caloric inputs, we observe the greater distance, speed, and heart rate are predicted, and vice versa. 
Moreover, Fig. \ref{fig:speed_hr_profile} shows the change in speed and heart rate with respect to time stamps. The blue line represents the original workout record, while the dotted lines represent the recommended speed and heart rate according to various input target caloric expenditures. We observe that the model can properly predict the fluctuation of speed and heart rate with relatively low MAE. By comparing three different predicted workout speed and heart rate sequences, the system can properly predict the correlation between distance, calories, and heart rate.

\begin{table}[t]
\centering
\begin{tabular}{llll} \toprule
\multicolumn{4}{c}{\textbf{Original workout record}}                                       \\ \midrule
\textbf{Calories} & \textbf{Distance} & \textbf{Speed AVG.} & \textbf{Heart Rate AVG.} \\ \midrule
592kcal           & 6.2km             & 8.8km/h             & 149bpm                   \\ \midrule
\multicolumn{4}{c}{\textbf{Recommended workout}}                                       \\ \midrule
\textbf{Calories} & \textbf{Distance} & \textbf{Speed AVG.} & \textbf{Heart Rate AVG.} \\ \midrule
474 kcal          & 5.97km            & 8.46km/h            & 136 bpm                  \\ 
\textbf{592 kcal} & \textbf{6.18km}   & \textbf{8.64km/h}   & \textbf{142 bpm}         \\ 
651 kcal          & 6.28km            & 8.69km/h            & 145bpm                   \\ \bottomrule
\end{tabular}
\caption{Workout profile Table}
\label{table:workout_profile_table}
\end{table}

Our $(P^{3}FitRec)$ has a high degree of freedom in choosing sport type, route, and target calories, and the personalized recommendations of speed and heart rate change dynamically according to the workout route. Therefore, the system provides users with greater flexibility to plan and predict exercises, and to modify the speed and pace during exercises.

The proposed models, datasets and experimental data will be made available at the project repository for further extension and reproducibility studies. \protect\footnote{\url{https://github.com/BasemSuleiman/Personalized\_Intelligent\_Fitness\_Recommender}.}.

\section{Conclusion and Future Work}
\label{sec:conclusion}

In this paper, we propose $(P^{3}FitRec)$, a Multi-layer MLP and Multi-layer Bi-LSTM based framework that provides personalized fitness recommendations with privacy preservation. Our $(P^{3}FitRec)$ employs Tensor Decomposition to infer entity embeddings from historical workout data and has achieved satisfactory results on predicting workout distance, workout speed sequence, and workout heart rate sequence. We demonstrate that personalized fitness recommendations can be achieved using minimum identity information from the users.

For further studies, we propose extending the work in three aspects. First, the cold-start problem is an interesting topic in building recommender systems, for which in our case is on new sport types and new users. Users playing new sports will have to exercise without personalized recommendation and contribute an adequate amount of workout data before a model can be trained to cover these sport types. Likewise, for new users who have absolutely no workout history, it is a challenge to learn user embeddings using the Tensor Decomposition method. Moreover, our $(P^{3}FitRec)$ is built on a two-model pipeline, instead of an end-to-end model structure to realize the two related tasks respectively. Though the models share the same contextual input features and the prediction of the first model feeds into the second model, one can argue that the accuracy of the application can be further improved if an end-to-end structure is applied, such as an encoder-decoder structure. Lastly, the Transformer architecture \cite{transformer} has become dominant in sequence prediction tasks, especially in the field of natural language processing. Although our experiment on the attention mechanism shows little performance enhancement, it is possible to improve the performance by trying the Transformer architecture.

%
%
\bibliographystyle{splncs04}
\bibliography{bibliography.bib}

\newpage
\appendix

\section{Appendices}
\label{sec:appendices}

\verb|Entity Embedding T-SNE plot|

\begin{figure}[ht]
\centering
\includegraphics[width=\textwidth]{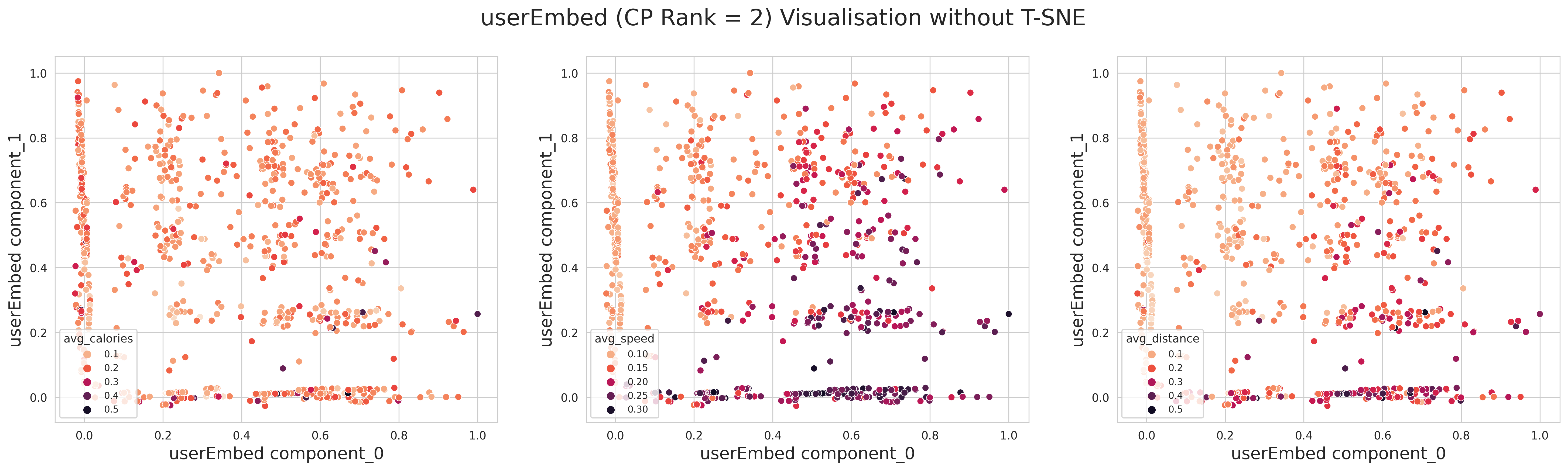}
\caption{Rank 2 User Embedding}
\label{fig:rank_2_user_embedding}

\end{figure}

\begin{figure}[ht]
\centering
\includegraphics[width=\textwidth]{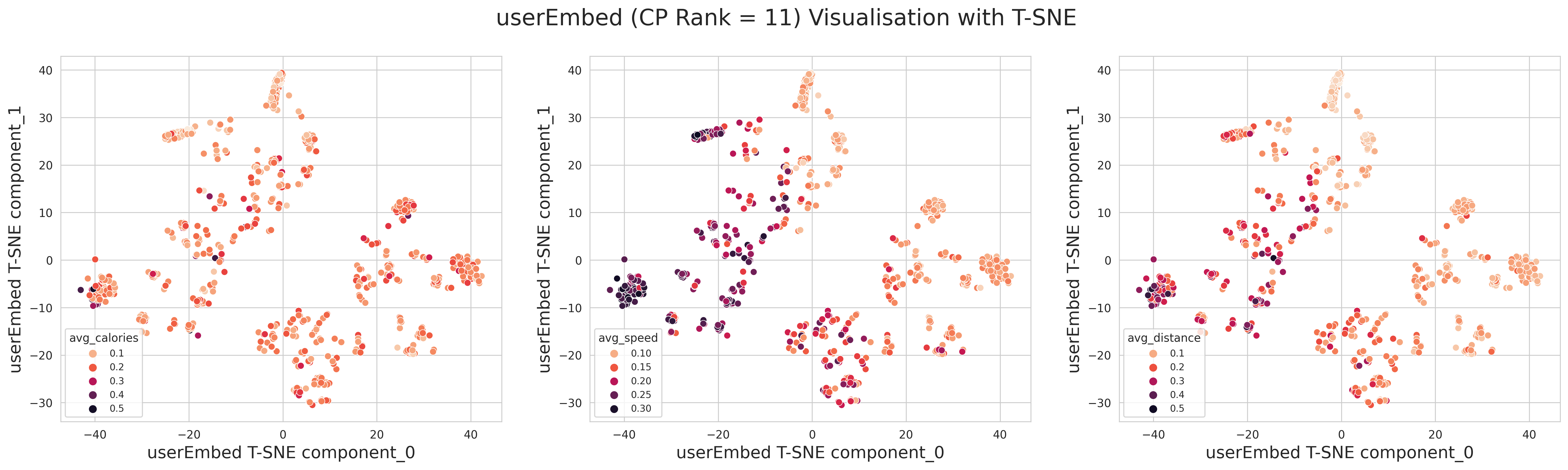}
\caption{Rank 11 User Embedding}
\label{fig:rank_11_user_embedding}

\end{figure}

\begin{figure}[ht]
\centering
\includegraphics[width=\textwidth]{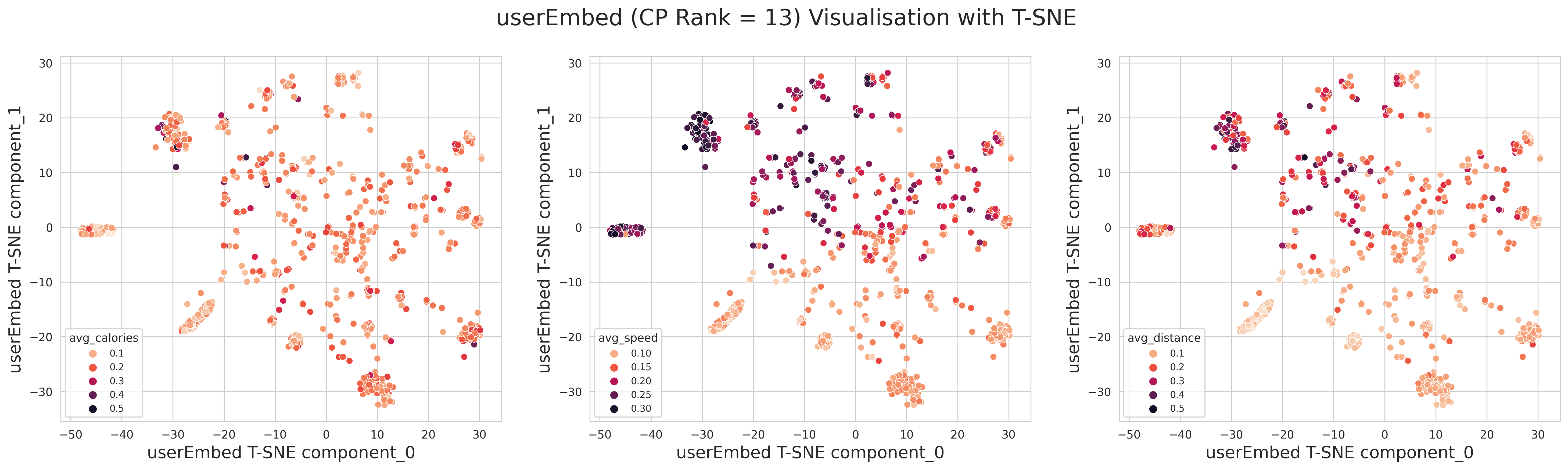}
\caption{Rank 13 User Embedding}
\label{fig:rank_13_user_embedding}

\end{figure}

\begin{figure}[ht]
\centering
\includegraphics[width=\textwidth]{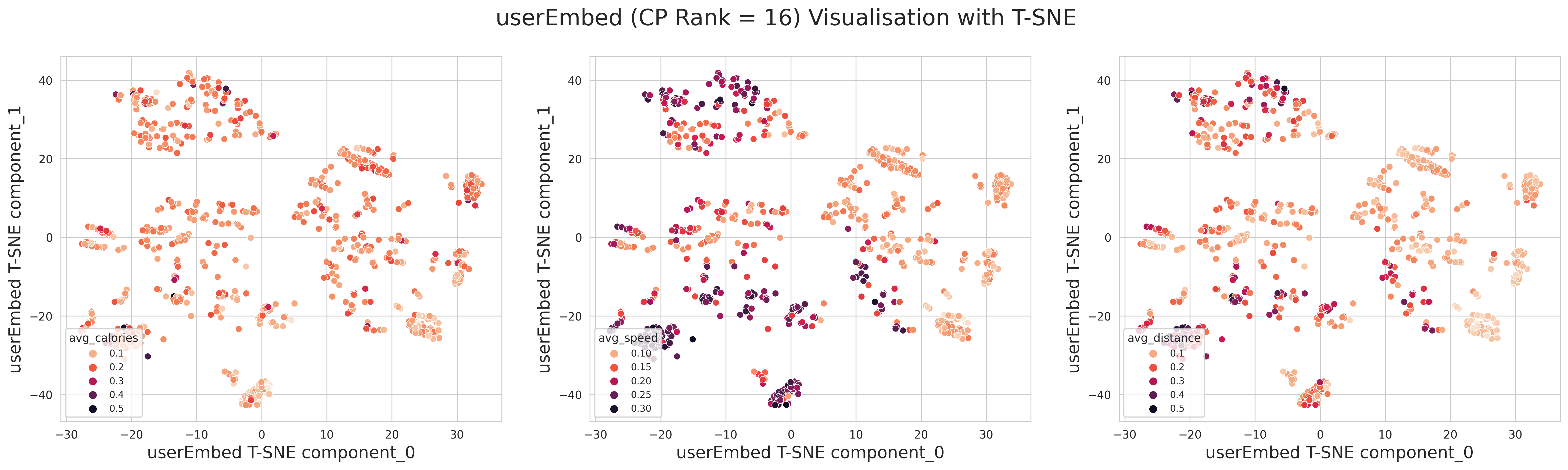}
\caption{Rank 16 User Embedding}
\label{fig:rank_16_user_embedding}
\end{figure}

\begin{figure}[ht]
\centering
\includegraphics[width=\textwidth]{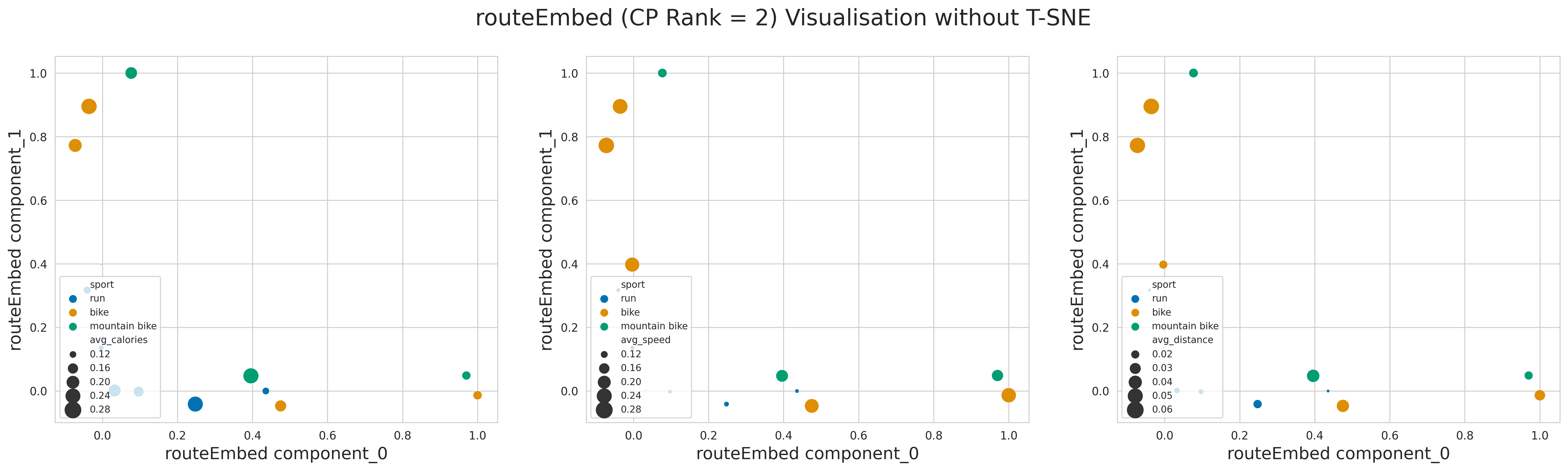}
\caption{Rank 2 Route Embedding}
\label{fig:rank_2_route_embedding}
\end{figure}

\begin{figure}[ht]
\centering
\includegraphics[width=\textwidth]{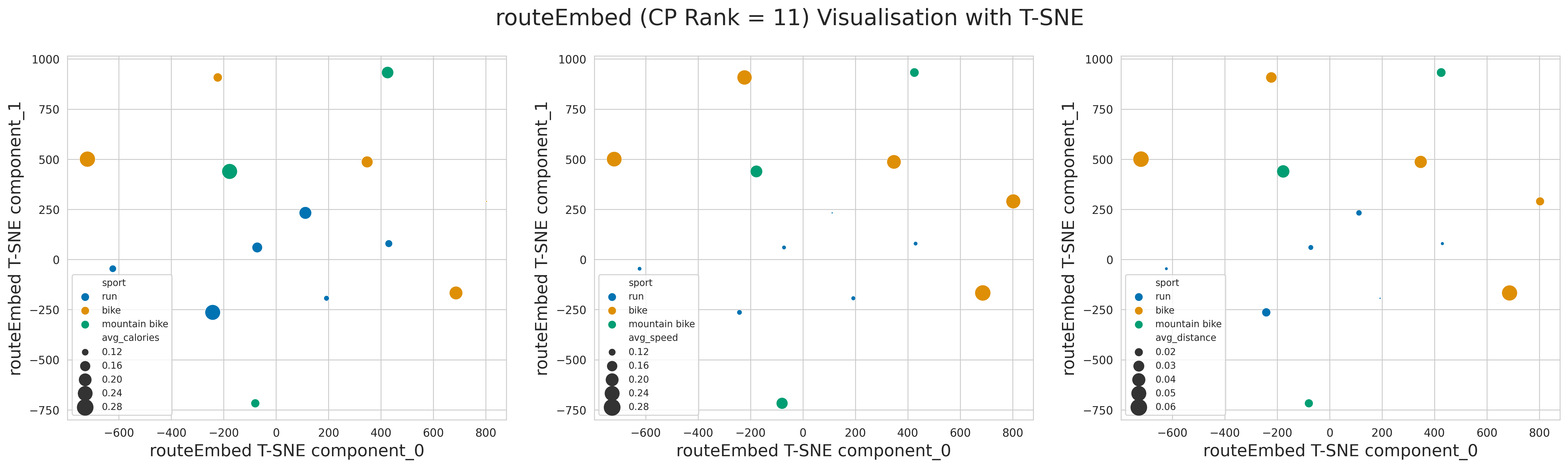}
\caption{Rank 11 Route Embedding}
\label{fig:rank_11_route_embedding}
\end{figure}

\begin{figure}[ht]
\centering
\includegraphics[width=\textwidth]{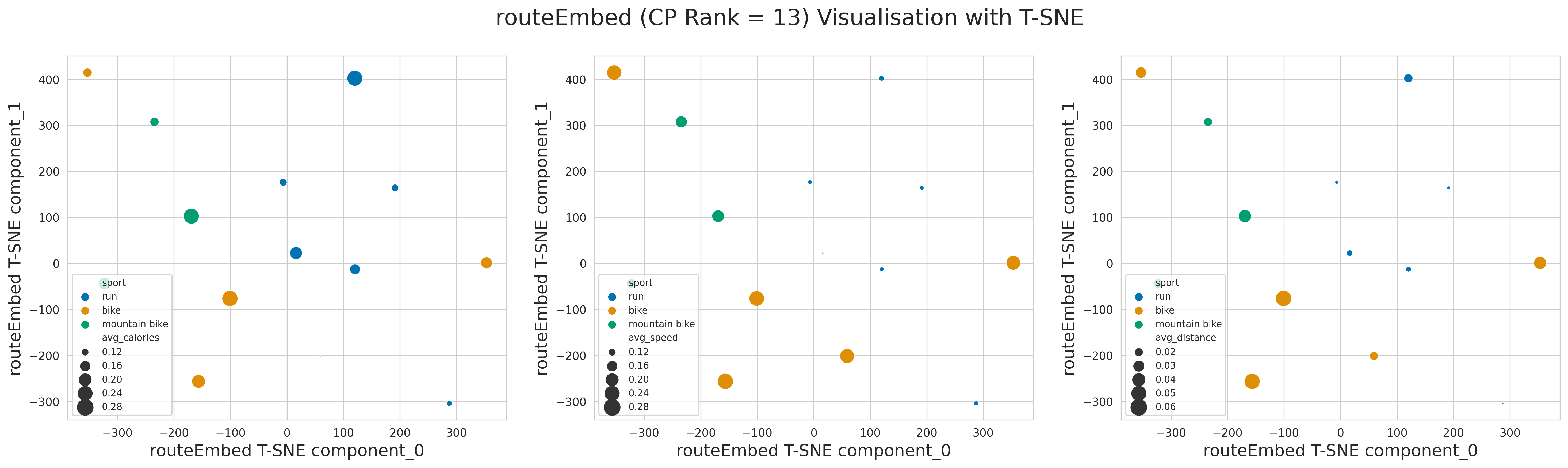}
\caption{Rank 13 Route Embedding}
\label{fig:rank_13_route_embedding}
\end{figure}

\begin{figure}[ht]
\centering
\includegraphics[width=\textwidth]{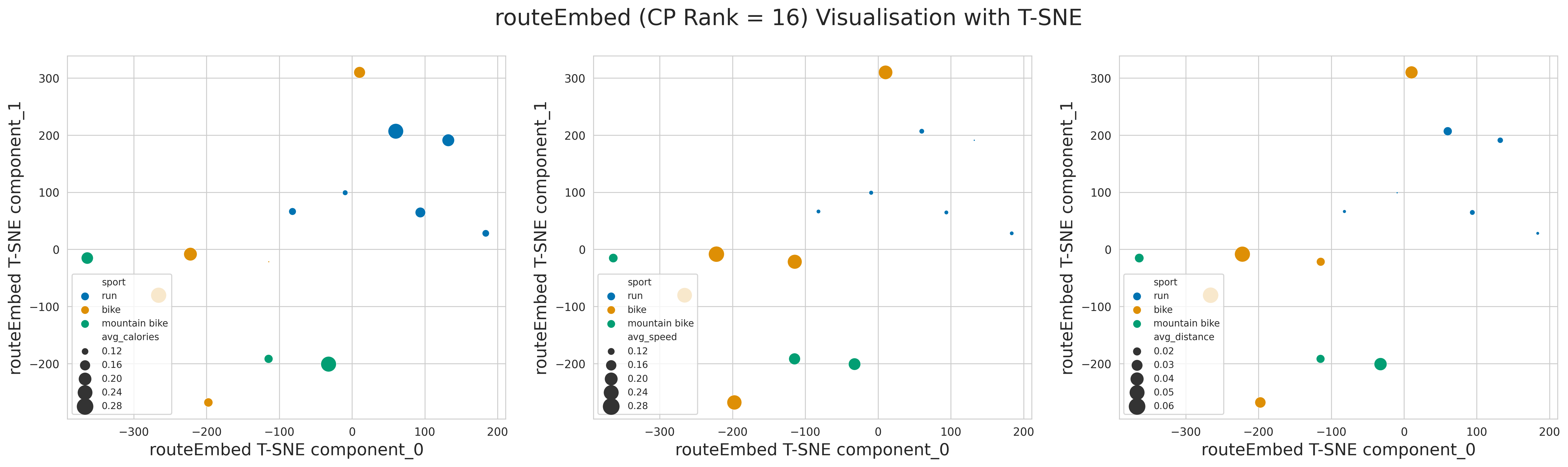}
\caption{Rank 16 Route Embedding}
\label{fig:rank_1_route_embedding}
\end{figure}

\end{document}